%% file: main.tex
\pdfoutput=1

\documentclass[11pt]{article}

\usepackage[]{naacl2021}

\usepackage{times}
\usepackage{latexsym}

\usepackage[T1]{fontenc}

\usepackage[utf8]{inputenc}

\usepackage{microtype}
\usepackage{latexsym,amsmath,amssymb,amsthm}

\usepackage{cleveref}

\usepackage{mystyle}
\usepackage{symbol}
\usepackage{mlsymbols}

\usepackage{paralist}
\usepackage{sistyle}
\SIthousandsep{,}
\usepackage{tikz}

\input{common}

\title{\theTitle}

\author{Yichu Zhou \\
  School of Computing \\
  University of Utah \\
  \texttt{flyaway@cs.utah.edu} \\\And
  Vivek Srikumar \\
  School of Computing \\
  University of Utah \\
  \texttt{svivek@cs.utah.edu} \\}

\begin{document}
\maketitle

\input{abstract}

\input{introduction}

\input{quality}
\input{approximation}
\input{reptask}
\input{analysis}

\input{related}
\input{conclusion}

\input{acknowledgments}

\bibliography{anthology,custom}
\bibliographystyle{acl_natbib}

\appendix
\input{appendix}
\input{details}

\end{document}

%% file: common.tex
\newcommand{\elmo}{ELMo\xspace}
\newcommand{\bertBaseCased}{BERT$_{\text{base,cased}}$\xspace}
\newcommand{\bertLargeCased}{BERT$_{\text{large,cased}}$\xspace}
\newcommand{\robertaBase}{RoBERTa$_{\text{base}}$\xspace}
\newcommand{\robertaLarge}{RoBERTa$_{\text{large}}$\xspace}

\newcommand{\probename}{{\sc DirectProbe}\xspace}

\newcommand{\theTitle}{\vs{Make up a paper title}}
\renewcommand{\theTitle}{Probing: Classification Is Not Necessary}
\renewcommand{\theTitle}{Anticipate Representation Performance without Classification}
\renewcommand{\theTitle}{Empirical Studies in Classifier-free Evaluation of
Representations\vs{This is the old EMNLP submission title}}
\renewcommand{\theTitle}{Skip Classifiers: Probe Representations Directly and Independently}
\renewcommand{\theTitle}{\probename: Studying Representations without Classifiers}


%% file: abstract.tex
\begin{abstract}

    Understanding how linguistic structure is encoded in
    contextualized embedding could help explain their
    impressive performance across NLP\@. Existing approaches
    for probing them usually call for training classifiers
    and use the accuracy, mutual information, or
    complexity as a proxy for the representation's goodness.
    In this work, we argue that doing so can be unreliable
    because different representations may need different
    classifiers. We develop a heuristic, \emph{\probename}, that
    directly studies the geometry of a representation by
    building upon the notion of a \emph{version space} for a task.
    Experiments with several linguistic tasks and
    contextualized embeddings show that, even without
    training classifiers, \probename can shine light into
    how an embedding space represents labels, and also
    anticipate classifier performance for the
    representation.

\end{abstract}


%% file: introduction.tex
\section{Introduction}\label{sec:introduction}

Distributed representations of
words~\cite[e.g.,][]{peters-etal-2018-deep,devlin-etal-2019-bert}
have propelled the state-of-the-art across NLP to new heights.
Recently, there is much interest in probing 
these opaque representations to understand the information they bear~\cite[e.g.,][]{kovaleva-etal-2019-revealing,conneau-etal-2018-cram,jawahar-etal-2019-bert}.
The most commonly used
strategy calls for training classifiers on them to predict linguistic
properties such as syntax, or cognitive skills like
numeracy~\cite[\eg][]{kassner-schutze-2020-negated,perone2018evaluation,yaghoobzadeh-etal-2019-probing,krasnowska-kieras-wroblewska-2019-empirical,wallace-etal-2019-nlp,pruksachatkun-etal-2020-intermediate}.
Using these classifiers, criteria such as 
accuracy or model complexity are used to evaluate the
representation quality for the
task~\cite[\eg][]{goodwin-etal-2020-probing,pimentel-etal-2020-pareto,michael2020asking}.

Such classifier-based probes are undoubtedly useful to estimate
a representation's quality for a task.
However, their ability to reveal the information in a representation is 
occluded by numerous factors, such as the choice of the optimizer and
the initialization used to train the classifiers.
For example, in our experiments using the task of
preposition supersense
prediction~\cite{schneider-etal-2018-comprehensive}, we
found that the accuracies across different training runs of
the same classifier can vary by as much as $\sim 8\%$!
(Detailed results can be found in \Cref{sec:cls}.)

Indeed, the very choice of a classifier influences our estimate of the quality
of a representation. For example, one representation 
may achieve the best classification accuracy with a linear model, whereas another
may demand a multi-layer perceptron for its
non-linear decision boundaries.
Of course, enumerating every possible classifier for a task is untenable.
A common compromise involves using linear
classifiers to probe representations
~\cite{alain2016understanding,kulmizev-etal-2020-neural}, but doing so may
mischaracterize representations that need non-linear separators.
Some work recognizes this problem~\cite{hewitt-liang-2019-designing} and proposes to report
probing results for at least logistic regression and
a multi-layer perceptron~\cite{eger-etal-2019-pitfalls}, or to compare the learning
curves between multiple controls~\cite{talmor-etal-2020-olmpics}.
However, the success of these methods still depends on
the choices of classifiers.

In this paper, we pose the question: \emph{Can we evaluate the quality of a
  representation for an NLP task directly without relying on classifiers as a
  proxy?}

Our approach is driven by a characterization of not one, but \emph{all} decision
boundaries in a representation that are consistent with a training set for a
task. This set of consistent (or approximately consistent) classifiers
constitutes the \emph{version space} for the task~\cite{mitchell1982generalization}, and
includes both simple (\eg, linear) and complex (\eg, non-linear) classifiers for
the task.
However, perfectly characterizing the version space for a problem presents
computational challenges. To develop an approximation, we note that any decision
boundary partitions the underlying feature space into contiguous regions
associated with labels. We present a heuristic approach
called \emph{\probename}, which builds upon hierarchical
clustering to identify such regions for a given task and embedding.

The resulting partitions allow us to directly probe the embeddings via their
geometric properties. For example, distances between these regions correlate
with the difficulty of learning with the representation: larger distances
between regions of different labels indicates that there are more consistent
separators between them, and imply easier learning, and better generalization
of classifiers.
Further, by assigning test points to their closest partitions, we have a
parameter-free classifier as a side effect, which can help benchmark
representations without committing to 
a specific family of classifiers (\eg, linear) as probes.

Our experiments study five different NLP tasks that involve syntactic and
semantic phenomena. 
We show that our approach allows us to ascertain, without training a classifier,
\begin{inparaenum}[(a)]
\item if a representation admits a linear separator for a dataset,
\item how different layers of BERT differ in their representations for a task,
\item which labels for a task are more confusable,
\item the expected performance of the best classifier for the task, and 
\item the impact of fine-tuning.
\end{inparaenum}

In summary, the contributions of this work are:
\begin{enumerate}
\item We point out that training classifiers as probes is not
  reliable, and instead, we should directly analyze the structure of
  a representation space.
\item We formalize the problem of evaluating representations via the notion of
  version spaces and introduce \probename, a heuristic method to approximate it
  directly which does not involve training classifiers.
\item Via experiments, we show that our approach can help identify how good a
  given representation will be for a prediction task.\footnote{\probename can be
    downloaded from \url{https://github.com/utahnlp/DirectProbe}.}
\end{enumerate}


%% file: quality.tex
\section{Representations and Learning}\label{sec:quality}
In this section, we will first briefly review the relationship
between representations and model learning. Then, we will introduce the notion of 
$\epsilon$-version spaces to characterize representation quality.

\subsection{Two Sub-tasks of a Learning Problem}\label{sec:subtasks}

The problem of training a predictor for a task can be divided into two
sub-problems: (a) representing data, and (b) learning a
predictor~\cite{bengio2013representation}. The former involves
transforming input objects $x$---words, pairs of words,
sentences, etc.---into a representation $E(x)\in\mathcal{R}^n$ that
provides features for the latter.  Model learning builds
a classifier $h$ over $E(x)$ to
make a prediction, denoted by the function composition
$h(E(x))$.

\begin{figure}
    \centering
    \includegraphics[width=0.5\textwidth]{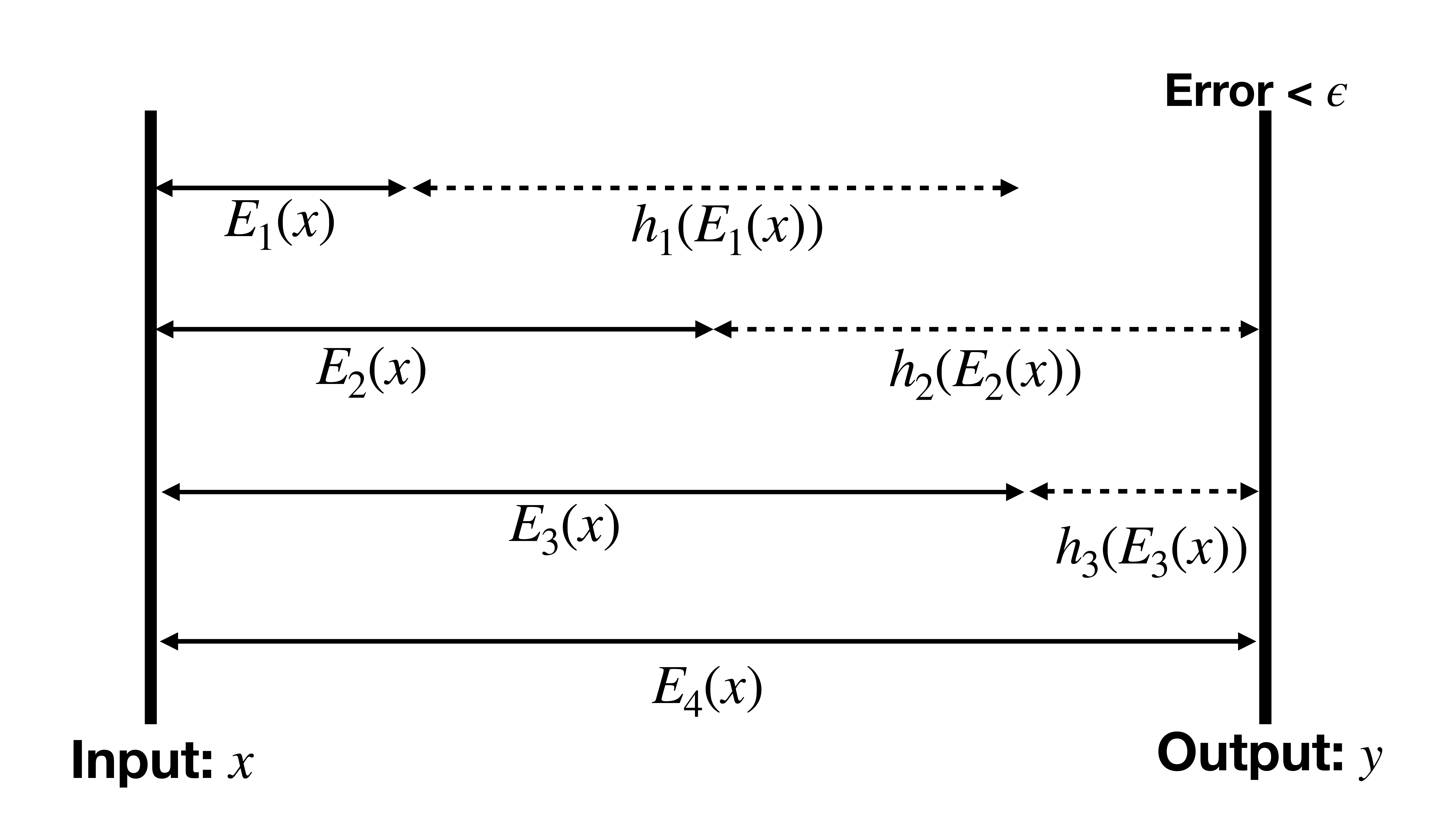}
    \caption{An illustration of the total work in a learning problem, 
    visualized as the distance traveled from the input bar
    to the output bar. Different representations $E_i$ and
    different classifiers $h_i$ provide different effort.
    }\label{fig:subworks}
\end{figure}

\Cref{fig:subworks} illustrates the two sub-tasks and their
roles in figuratively ``transporting'' an input $x$ towards a prediction with a
probability of error below a small
$\epsilon$. For the representation $E_1$, the best classifier $h_1$ falls short
of this error requirement. The representation $E_4$ does not
need an additional classifier because it is identical to the label. The
representations $E_2$ and $E_3$ both admit classifiers $h_2$ and $h_3$ that meet
the error threshold. Further, note,  that $E_3$ leaves less work
for the classifier than $E_2$, suggesting that it is a better representation as
far as this task is concerned.

This illustration gives us the guiding principle for this work\footnote{In
  addition to performance and complexity of the best classifier, other aspects
  such as sample complexity, the stability of learning, etc are also
  important. We do not consider them in this work: these aspects are related to
  optimization and learnability, and more closely tied to classifier-based probes.}:
\begin{quote}
  The quality of a representation $E$ for a task is a function of both the
  performance and the complexity of the \emph{best} classifier $h$ over that
  representation. 
\end{quote}
Two observations follow from the above discussion. First, we cannot enumerate
every possible classifier to find the best one. Other recent work, such as that of \citet{xia-etal-2020-predicting}
make a similar point. Instead, we need to resort to an
approximation to evaluate representations. Second, trivially, the best
representation for a task is identical to an accurate classifier; in the
illustration in \Cref{fig:subworks}, this is represented by
$E_4$. However, such a representation is over-specialized to one task. In contrast,
learned representations like BERT promise \emph{task-independent}
representations that support accurate classifiers.



\subsection{$\epsilon$-Version Spaces}\label{sec:signal}

Given a classification task, we seek to disentangle the evaluation of a representation
$E$ from the classifiers $h$ that are trained over it.
To do so, the first step is to characterize all classifiers supported
by a representation.

Classifiers are trained to find a hypothesis (\ie, a classifier) that is
consistent with a training set. A representation 
$E$ admits a set of such hypotheses, and a learner
chooses one of them.  Consider the top-left example 
of \Cref{fig:lines}. There are many classifiers that separate the two classes;
the figure shows 
two linear ($h_1$ and $h_2$) and one non-linear ($h_3$)
example.
Given a set $\mathcal{H}$ of classifiers of interest, the subset of classifiers that are consistent with a given
dataset represents the version space with respect to
$\mathcal{H}$~\cite{mitchell1982generalization}. 
To account for errors or noise in data, we
define an $\epsilon$-version space: \textit{the set of
hypothesis that can achieve less than $\epsilon$ error on a
given dataset}.

Let us formalize this definition. Suppose $\mathcal{H}$
represents the whole hypothesis space consisting of all
possible classifiers $h$ of interest. The $\epsilon$-version space
$V_{\epsilon}(\mathcal{H},E,D)$ expressed by a
representation $E$ for a labeled dataset $D$ is defined as:
\begin{equation}
{
    V_{\epsilon}\left(\mathcal{H}, E, D\right) \triangleq \left\{ h \in
    \mathcal{H} \mid err(h, E, D) \leq \epsilon \right\}
}
\end{equation}
where $err$ represents training error. 

Note that the $\epsilon$-version space
$V_{\epsilon}(\mathcal{H},E,D)$ is only a set of functions
and does not involve any learning.
However, understanding a representation 
requires examining its $\epsilon$-version space---a larger one would allow
for easier learning.

In previous
work, the quality of a representation $E$ for a task represented by a dataset
$D$ is measured via properties of a specific 
$h\in V_{\epsilon}(\mathcal{H},E,D)$, typically a linear probe.  Commonly
measured properties include
generalization
error~\cite{kim-etal-2019-probing},
minimum description length of labels 
~\cite{voita-titov-2020-information} and
complexity~\cite{whitney2020evaluating}. 
%
Instead, we seek to directly evaluate the $\epsilon$-version space of a
representation for a task, without committing to a restricted set of probe classifiers.


%% file: approximation.tex
\section{Approximating an $\epsilon$-Version Space}\label{sec:dectecting}

Although the notion of an $\epsilon$-version space is well defined, finding it
for a given representation and task is impossible because it involves
enumerating all possible classifiers.
In this section, we will describe a heuristic
to approximate the $\epsilon$-version space. We
call this approach as \emph{\probename}.

\subsection{Piecewise Linear Approximation}\label{sec:piecewise}

Each classifier in $V_{\epsilon}(\mathcal{H},E,D)$ is a decision boundary in the
representation space that separates examples with different labels
(see~\Cref{fig:lines}, left). 
The decisions of a classifier can be mimicked by a set of
piecewise linear functions.  \Cref{fig:lines} (middle) shows two examples.
At the top is the simple case with linearly separable points.  
At the bottom is a more complicated case with a circular
separator. The set of the piecewise linear function that matches
its decisions needs at least three lines. 

\input{lines}

The ideal piecewise linear separator partitions training points into groups,
each of which contains points with exactly one label. These groups can be seen
as defining convex regions in the embedding space (see~\Cref{fig:lines}, left).
Any classifier in $V_{\epsilon}(\mathcal{H},E,D)$ must cross the regions
\emph{between} the groups with different labels; these are the regions that
separate labels from each other, as shown in the gray areas in \Cref{fig:lines}
(middle). Inspired by this, we posit that these regions between groups with
different labels, and indeed the partitions themselves, offer insight into
$V_{\epsilon}(\mathcal{H},E,D)$.

\subsection{Partitioning Training Data}\label{sec:clustering}

Although finding the set of all decision boundaries remain
hard, finding the regions between convex groups that these
piecewise linear functions splits the data into is less so.
Grouping data points in this fashion is related to a
well-studied problem, namely
clustering, and several recent works have looked at clustering of contextualized
representations~\cite{reimers-etal-2019-classification,aharoni-goldberg-2020-unsupervised,gupta2020clustering}.
In this work, we have a new clustering problem
with the
following
criteria:
\begin{inparaenum}[(i)]
\item \textit{All points in a group have the same label.} We need to ensure we are mimicking the decision boundaries.
    \item \textit{There are no overlaps between the convex hulls of each group.} If convex hulls of two groups do not overlap, there must exist a line that can separates them, as guaranteed by the hyperplane separation theorem. 
    \item \textit{Minimize the number of total groups.} Otherwise, a simple solution is that each data point becomes a group by itself.
\end{inparaenum}


Note that the criteria do not require all points of one label to be grouped
together. For example, in~\Cref{fig:lines} (bottom right),
points of the  circle (\ie, $\circ$) class are in three
subgroups.

To summarize what we have so far: we transformed the problem
of finding the $\epsilon$-version space into a clustering
problem with specific criteria. Next, let us see a heuristic
for partitioning the training set into clusters based on
these criteria.

\subsection{A Heuristic for Clustering}\label{sec:heuristic}

To find clusters as described in \Cref{sec:clustering}, we
define a simple bottom-up heuristic clustering strategy that forms the basis of \probename
(\Cref{alg:simple-bottom-up}). In the beginning,
each example $x_i$ with label $y_i$ in a training set $D$ is
a cluster $C_i$ by itself. In each iteration, we select the
closest pair of clusters with the same label and merge them
(lines 4, 5). If the convex hull of the new cluster does not
overlap with all other clusters\footnote{Note that ``all
other clusters'' here means the other clusters with
different labels. There is no need to prohibit overlap
between clusters with the same label since they might be
merged in the next iterations.}, we keep this new cluster
(line 9). Otherwise, we flag this pair (line 7) and choose
the next closest pair of clusters. We repeat these steps
till no more clusters can be merged.

\begin{algorithm}
    \small
    \caption{\probename: Bottom-up Clustering}
    \begin{algorithmic}[1]
        \Require{A dataset $D$ with labeled examples $(x_i, y_i)$.}
        \Ensure{A set of clusters of points $\mathcal{C}$.}
    
        \State{Initialize $C_i$ as the cluster for each
        $(x_i,y_i)\in D$}
        \State{$\mathcal{C}=\{C_0, C_1, \cdots\}, B=\emptyset$}
        \Repeat
            \State{Select the closest pair $(C_i,C_j)\in
            \mathcal{C}$ which have the same label and is
            not flagged in $B$.
            }
            \State{$ S \leftarrow C_i \cup C_j$}
            \If{convex hull of $S$ overlaps with other elements of $\mathcal{C}$}
                \State{$B\leftarrow B \cup \{(C_i,C_j)\}$}
            \Else
                \State{Update $\mathcal{C}$ by removing $C_i, C_j$ and adding $S$}
            \EndIf
        \Until{no more pairs can be merged}
        \State{\Return{$\mathcal{C}$}}
    \end{algorithmic}\label{alg:simple-bottom-up}
\end{algorithm}

We define the distance between clusters $C_i$ and $C_j$ as the Euclidean distance between their centroids. 
%
Although \Cref{alg:simple-bottom-up} does not guarantee the
minimization criterion in~\Cref{sec:clustering} since it is greedy heuristic, we will see in
our experiments that, in practice, it works well. 
 
\paragraph{Overlaps Between Convex Hulls}
A key point of~\Cref{alg:simple-bottom-up} is checking if
the convex hulls of two sets of points overlap (line 6).  Suppose we have two
sets $C=\{x_1^C,\ldots,x_n^C\}$ and
$C^{\prime}=\{x_1^{C^{\prime}},\ldots,x_m^{C^{\prime}}\}$. We can restate this as the problem of checking if there is some vector $w\in\Re^n$ and a number $b\in\Re$ such that:
\begin{equation}
  \begin{aligned}
    \forall x_i^C \in C,&~\quad w^{\top}E(x_i^{C}) + b \geq 1,  \\
    \forall x_j^{C^\prime} \in {C^\prime},&~\quad w^{\top}E(x_j^{C^{\prime}})  + b \leq -1.
  \end{aligned}
\end{equation}
where $E$ is the representation under investigation.

We can state this problem as a linear program that checks for feasibility of the system of inequalities.
%
%
If the LP problem is feasible, there must exist a separator
between the two sets of points, and they do not overlap. In
our implementation, we use the Gurobi
optimizer~\cite{gurobi} to solve the linear
programs.\footnote{\Cref{alg:simple-bottom-up} can be made
faster by avoiding unnecessary calls to the
solver. \Cref{sec:detailed-algorithm} gives a 
detailed description of these techniques, which are also incorporated in the
code release.}

\subsection{Noise: Controlling $\epsilon$}\label{sec:noise}

Clusters with only a small number of points could be treated as noise.
Geometrically, a point in the neighborhood of other points with different
labels could be thought of as noise.  Other clusters can not
merge with it because of the no-overlap constraint. As a result, such
clusters will have only a few points (one or two in practice).
If we want
zero error rate on the training data, we can keep these noise points; if we allow
a small error rate $\epsilon$, then we can remove these noise clusters. In our
experiments, for simplicity, we keep all clusters.




%% file: lines.tex
\begin{figure}
  \centering
  \begin{tikzpicture}

    \node at (0.3,2.8) {$\circ$};
    \node at (0.6,3) {$\circ$};
    \node at (1,3) {$\circ$};
    \node at (0.3,3.3) {$\circ$};
    \node at (1.39, 2.95) {\tiny $\blacksquare$};
    \node at (1.85,2.75) {\tiny $\blacksquare$};
    \node at (1.85,3.15) {\tiny $\blacksquare$};

    \draw [thick] (1.2, 4) -- (1.2,2.2);
    \draw [thick] (1.6, 4) -- (0.9,2.2);
    \draw [thick] (0.5, 3.7) .. controls (1.33,3) ..
    (0.6,2.5);

    \node at (1.2,4.2) {$h_1$};
    \node at (1.8,4.1) {$h_2$};
    \node at (0.4,4.0) {$h_3$};

    \begin{scope}[xshift=2.5cm, yshift=2.4cm]
      \path [fill=gray!40] (1.05,0.6) rectangle (1.3,1.0);
      \node at (0.3,0.6) {$\circ$};
      \node at (0.6,0.8) {$\circ$};
      \node at (1,0.8) {$\circ$};
      \node at (0.3,1.1) {$\circ$};
      \node at (1.39, 0.75) {\tiny $\blacksquare$};
      \node at (1.85, 0.55) {\tiny $\blacksquare$};
      \node at (1.85, 0.95) {\tiny $\blacksquare$};
      \draw [dashed,thick] (1.2, 1.8) -- (1.2,0);
    \end{scope}

    \begin{scope}[xshift=5cm, yshift=2.4cm]
      \node (tc0) at (0.3,0.6) {$\circ$};
      \node (tc1) at (0.6,0.8) {$\circ$};
      \node (tc2) at (1,0.8) {$\circ$};
      \node (tc3) at (0.3,1.1) {$\circ$};
      \node (ts0) at (1.39, 0.75) {\tiny $\blacksquare$};
      \node (ts1) at (1.85, 0.55) {\tiny $\blacksquare$};
      \node (ts2) at (1.85, 0.95) {\tiny $\blacksquare$};

      \draw[solid] (tc0.center) -- (tc2.center) -- (tc3.center) -- (tc0.center);
      \draw[solid] (ts0.center) -- (ts1.center) -- (ts2.center) -- (ts0.center);
    \end{scope}

    \begin{scope}[xshift=-3cm, yshift=-2.4cm]
      \node at (4,3.1)  {\tiny $\blacksquare$};
      \node at (3.95,2.85)  {\tiny $\blacksquare$};
      \node at (4.15,2.9)  {\tiny $\blacksquare$};
      \draw [thick] (4,3) circle [radius=0.5];

      \node at(3.2,3) {$\circ$};
      \node at(3.4,2.471) {$\circ$};
      \node at(3.4,3.529) {$\circ$};
      \node at(3.7,2.258) {$\circ$};
      \node at(3.7,3.742) {$\circ$};
      \node at(4.2,2.225) {$\circ$};
      \node at(4.2,3.775) {$\circ$};
      \node at(4.6,2.471) {$\circ$};
      \node at(4.6,3.529) {$\circ$};
      \node at(4.8,3) {$\circ$};
    \end{scope}

    \begin{scope}[xshift=-0.5cm]
      \path [fill=gray!40] (4,0.6) circle [radius=0.70];
      \path [fill=white] (4,0.6) circle [radius=0.27];
      \node at (4,0.7)  {\tiny $\blacksquare$};
      \node at (3.95,0.45) {\tiny $\blacksquare$};
      \node at (4.15,0.5) {\tiny $\blacksquare$};

      \draw [dashed,thick] (3.7, 1.6) -- (5.2,-0.1);
      \draw [dashed,thick] (4.1, 1.6) -- (3.0,-0.2);
      \draw [dashed,thick] (3.0, 0.25) -- (5.2,0.25);

      \node at (3.2,0.6) {$\circ$};
      \node at (3.4,0.071) {$\circ$};
      \node at (3.4,1.129) {$\circ$};
      \node at (3.7,-0.142) {$\circ$};
      \node at (3.7,1.342) {$\circ$};
      \node at (4.2,-0.175) {$\circ$};
      \node at (4.2,1.375) {$\circ$};
      \node at (4.6,0.071) {$\circ$};
      \node at (4.6,1.129) {$\circ$};
      \node at (4.8,0.6) {$\circ$};
    \end{scope}

    \begin{scope}[xshift=2.5cm]
      \node (s0) at (4,0.7)  {\tiny $\blacksquare$};
      \node (s1) at (3.95,0.45) {\tiny $\blacksquare$};
      \node (s2) at (4.15,0.5) {\tiny $\blacksquare$};

      \node (c0) at (3.2,0.6) {$\circ$};
      \node (c1) at (3.4,0.071) {$\circ$};
      \node (c2) at (3.4,1.129) {$\circ$};
      \node (c3) at (3.7,-0.142) {$\circ$};
      \node (c4) at (3.7,1.342) {$\circ$};
      \node (c5) at (4.2,-0.175) {$\circ$};
      \node (c6) at (4.2,1.375) {$\circ$};
      \node (c7) at (4.6,0.071) {$\circ$};
      \node (c8) at (4.6,1.129) {$\circ$};
      \node (c9) at (4.8,0.6) {$\circ$};

      \draw[solid] (c0.center) -- (c4.center) -- (c2.center) -- (c0.center);
      \draw[solid] (c1.center) -- (c3.center) -- (c5.center) -- (c7.center) -- (c1.center);
      \draw[solid] (c6.center) -- (c8.center) -- (c9.center) -- (c6.center);
      \draw[solid] (s0.center) -- (s1.center) -- (s2.center) -- (s0.center);

    \end{scope}
    
  \end{tikzpicture}
  \caption{Using the piecewise linear functions to mimic
    decision boundaries on two different cases. The solid
    lines on the left are the real decision boundaries for
    a binary classification problem. The dashed lines in the
    middle are the piecewise linear functions used to mimic
    these decision boundaries. The gray area is the region
    that a separator must cross. The connected points in the right represent the convex regions  that the piecewise linear separators lead to.}\label{fig:lines}
\end{figure}
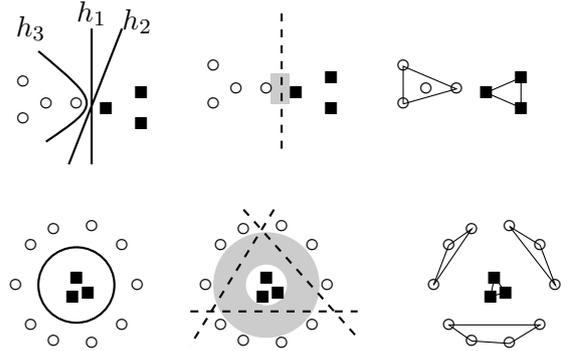


%% file: reptask.tex
\section{Representations, Tasks and Classifiers}\label{sec-task-rep}
Before looking at the analysis offered by the partitions obtained
via \emph{\probename} in~\Cref{sec:analysis}, let us first enumerate
the English NLP tasks and representations we will encounter. 

\subsection{Representations}
Our main experiments focus on \bertBaseCased, and we also show
additional analysis on other contextual
representations:
\elmo~\cite{peters-etal-2018-deep}\footnote{For simplicity,
we use the equally weighted three layers of \elmo in our
experiments.}, \bertLargeCased~\cite{devlin-etal-2019-bert},
\robertaBase and
\robertaLarge~\cite{liu2019roberta}. We refer the reader to
the \Cref{sec:rep-summarization} for further details about
these embeddings.
We use the average of subword embeddings as the
token vector for the representations that use
subwords. We use the original implementation of \elmo, and the HuggingFace library~\cite{wolf-etal-2020-transformers} for the others.

\subsection{Tasks}

We conduct our experiments on five NLP tasks that cover the
varied usages of word representations (token-based,
span-based, and token pairs) and include both
syntactic and semantic prediction problems. The \Cref{sec:task-summarization} has
more details about the tasks to help with replication.

\textbf{Preposition supersense disambiguation} represents a pair of
tasks, involving classifying a preposition's
\textbf{semantic role (SS-role)} and \textbf{semantic
function (SS-func)}.  Following the previous
work~\cite{liu-etal-2019-linguistic}, we only use the
single-token prepositions in the Streusle v4.2
corpus~\cite{schneider-etal-2018-comprehensive}.

\textbf{Part-of-speech tagging (POS)} is a token level
prediction task. We use the English portion of the parallel
universal dependencies
treebank~\cite[ud-pud][]{nivre2016universal}.

\textbf{Semantic relation (SR)} is the task of predicting the
semantic relation between pairs of nominals. We use the
dataset of semeval2010 task
8~\cite{hendrickx-etal-2010-semeval}. To represent the pair
of nominals, we concatenate their embeddings. Some nominals
could be spans instead of individual tokens, and we represent them via the
average embedding of the tokens in the span.

\textbf{Dependency relation (DEP)} is the task of predicting the syntactic
dependency relation between a token $w_{head}$ and its modifier $w_{mod}$.  We
use the universal dependency annotation of the English web
treebank~\cite{bies2012english}. As with semantic relations, to represent the pair of
tokens, we concatenate their embeddings.



\subsection{Classifier Accuracy}
The key starting point of this work is that restricting ourselves to linear
probes may be insufficient.
%
%
To validate the results of our analysis, we evaluate a large collection of
classifiers---from simple linear classifiers to two-layers neural networks---for
each task. For each one, we choose the best hyper-parameters using
cross-validation. From these classifiers, we find the best test
accuracy of each task and representation. All classifiers are trained with the
scikit-learn library~\cite{scikit-learn}. To reduce the impact of randomness,
we trained each classifier $10$ times with different initializations,
and report their average accuracy. The \Cref{sec:detailed-probing} summarizes the best classifiers
we found and their performance.


%% file: analysis.tex
\section{Experiments and Analysis}\label{sec:analysis}

\emph{\probename} 
helps partition an embedding space for a task, and thus
characterize  its $\epsilon$-version space.
Here, we will see that these clusters do indeed characterize various
linguistic properties of the representations  we consider.

\subsection{Number of Clusters}\label{sec:num-clusters}

\paragraph{The number of clusters is an indicator of the
linear separability of representations for a task.}
The best scenario is when the
number of clusters equals the number of labels. In this case,
examples with the same label are placed close enough by the
representation to form a cluster that is separable from
other clusters. A simple linear multi-class classifier can
fit well in this scenario. In contrast, if the number of
clusters is more than the number of labels, then some labels
are distributed across multiple clusters (as in
\Cref{fig:lines}, bottom). There must be a non-linear
decision boundary. Consequently, this scenario calls for a
more complex classifier, \eg, a multi-layer neural network.

In other words, using the clusters, and without training a classifier, we can
answer the question: \emph{can a linear classifier fit the training set for a
  task with a given representation?}

To validate our predictions, we use the training accuracy of
a linear SVM~\cite{chang2011libsvm} classifier.
If a linear SVM can perfectly fit ($100\%$
accuracy) a training set, then there exist linear decision
boundaries that separate the labels. \Cref{tb:linearity-pos}
shows the linearity experiments on the POS task,
which has 17 labels in total. All representations except
\robertaLarge have $17$ clusters, suggesting a linearly
separable space, which is confirmed by the SVM accuracy.
We conjecture that  this may be the reason  why linear models  usually work for
BERT-family 
models. Of course, linear separability does not mean the task is easy or that the
best classifier is a linear one. We found that, while
most representations we considered are linearly separable for most of
our tasks, the best classifier is not always linear. We refer the reader
to~\Cref{sec:detailed-probing} for the full results.

\begin{table}[]
\small
\centering
\begin{tabular}{lrr}
\toprule
    \multirow{2}{*}{Embedding}   & Linear SVM & \multirow{2}{*}{\#Clusters} \\
                & Training Accuracy &  \\ \midrule
\bertBaseCased  & 100        & 17        \\
\bertLargeCased & 100        & 17        \\
\robertaBase    & 100        & 17        \\
\robertaLarge   & 99.97      & 1487      \\
\elmo           & 100        & 17        \\ \bottomrule
\end{tabular}\caption{Linearity experiments on POS tagging
    task. Our POS tagging task has $17$ labels in
    total. Both linear SVM and the number of clusters
    suggest that \robertaLarge is non-linear while others
    are all linear, which means the best classifier for
    \robertaLarge
    is not a linear model. More details can be found in
    \Cref{sec:detailed-probing}.
    }\label{tb:linearity-pos}
\end{table}

\subsection{Distances between Clusters}\label{sec:min-dis}

As we mentioned in \Cref{sec:piecewise}, a learning process
seeks to find a decision boundary that separates clusters
with different labels.  Intuitively, a larger gap between
them would make it easier for a learner to find a suitable
hypothesis $h$ that generalizes better.


We use the distance between convex hulls of clusters as an indicator of
the size of these gaps. We note that the problem of computing the distance between convex hulls of
clusters is equivalent to finding the maximum margin separator between them. To
find the distance between two clusters,
we train a
linear SVM~\cite{chang2011libsvm} that separates them and compute its margin. The
distance we seek is twice the margin. For a given
representation, we are interested in the minimum
distance across all pairs of clusters with different labels.

\subsubsection{Minimum Distance of Different
Layers}\label{sec:min-dis-layers}

\paragraph{Higher layers usually have larger $\epsilon$-version spaces.}
Different layers of BERT play different roles when
encoding liguistic information~\cite{tenney-etal-2019-bert}.
To investigate the geometry of different layers of
BERT, we apply \probename to each
layer of \bertBaseCased for all five tasks. Then, we computed the 
minimum distances among all pairs of clusters with different
labels.
By comparing the minimum distances of
different layers, we answer the question: \emph{how
do different layers of BERT differ in their representations for
a task?}

\begin{figure}[]
    \centering
    \includegraphics[width=0.5\textwidth]{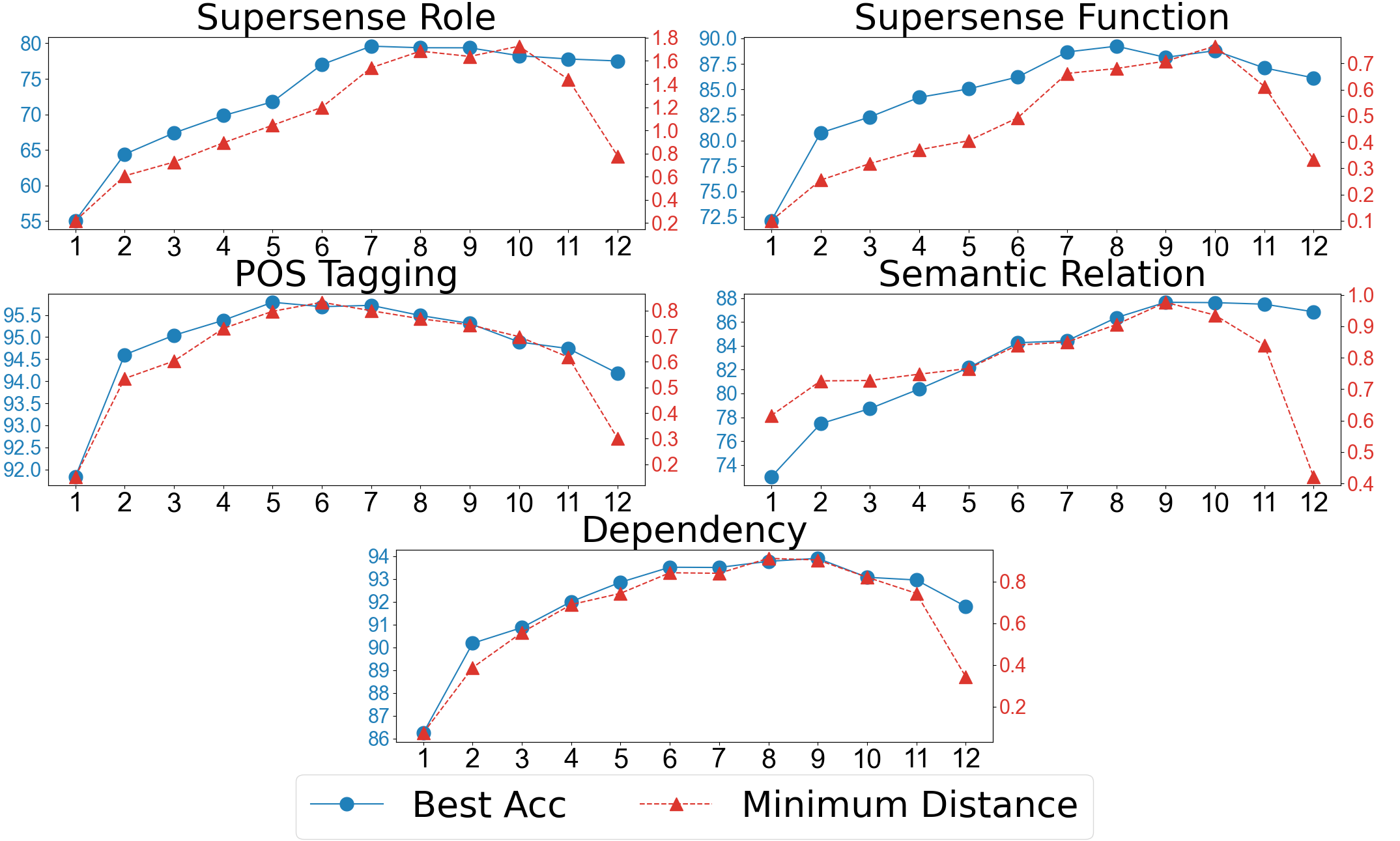}
    \caption{Here we juxtapose the minimum distances between
    clusters and the best classifier accuracy for all 12
    layers. The horizontal axis is the layer
    index of \bertBaseCased; the left vertical axis is the
    best classifier accuracy and the right vertical axis is the
    minimum distance between all pairs of
    clusters.}\label{fig:dis-layers}
\end{figure}

\Cref{fig:dis-layers} shows the results on all tasks. In
each subplot, the horizontal axis is the layer index. For each layer, the
blue circles (left vertical axis) is the best classifier accuracy, and the
red triangles (right vertical axis) is the minimum distance described above. We observe
that both best classifier accuracy and minimum distance show
similar trends across different layers: first increasing, then decreasing. It shows that minimum distance
correlates with the best performance for
an embedding space, though it is not a simple linear
relation. Another interesting observation is the decreasing
performance and minimum distance of higher layers,  which is
also corroborated by~\citet{ethayarajh-2019-contextual}
and~\citet{liu-etal-2019-linguistic}. 

\subsubsection{Impact of Fine-tuning}

\paragraph{Fine-tuning expands the $\epsilon$-version space.}
Past
work~\cite{peters-etal-2019-tune,arase-tsujii-2019-transfer,merchant-etal-2020-happens}
has shown that fine-tuning pre-trained models on a specific
task improves performance, and fine-tuning is now the \emph{de facto}
procedure for using contextualized embeddings. In this experiment, we try to
understand why fine-tuning can improve
performance.  Without
training classifiers, we answer the question: \emph{What changes in the
  embedding space after fine-tuning?}

We conduct the experiments described in
\Cref{sec:min-dis-layers} on the last layer of
\bertBaseCased before and after fine-tuning for all tasks.
Table~\ref{tb:min-dis-fine-tuning} shows the results. We
see that after fine-tuning, both the best classifier
accuracy and minimum distance show a big boost. It means
that fine-tuning pushes the clusters away from each other in the
representation space, which results in a larger
$\epsilon$-version space. As we discussed in
\Cref{sec:min-dis}, a larger $\epsilon$-version space
admits more good classifiers and allows for
better generalization. 

\begin{table}[]
\centering
\small
\begin{tabular}{llrr}
\toprule
Task &            & Min Distance &  Best Acc \\ \midrule
\multirow{2}{*}{SS-role}     & original   & 0.778  & 77.51                    \\
                         & fine-tuned & 4.231     & 81.62                    \\ \midrule
\multirow{2}{*}{SS-func}        & original   & 0.333    & 86.13                     \\
                         & fine-tuned & 2.686      & 88.4                     \\ \midrule
\multirow{2}{*}{POS}        & original  &  0.301  & 93.59                    \\
                         & fine-tuned &     0.7696 & 95.95                     \\ \midrule
\multirow{2}{*}{SR}     & original   &   0.421 & 86.85                    \\
                         & fine-tuned &      4.734 & 90.03                    \\ \midrule
\multirow{2}{*}{DEP} & original   &   0.345 & 91.52                   \\
                            & fine-tuned &    1.075 & 94.82 \\ \bottomrule
\end{tabular}
\caption{The best performance and the minimum distances between
    all pairs of clusters of the last layer of
    \bertBaseCased before and after
    fine-tuning.}\label{tb:min-dis-fine-tuning}
\end{table}

\subsubsection{Label Confusion}

\paragraph{Small distances between clusters can confuse a classifier.}
By comparing the distances between clusters, we can answer
the question: \emph{Which labels for a task are more
confusable?}

We compute the distances between all the pairs of labels
based on the last layer of \bertBaseCased.\footnote{For all
tasks, \bertBaseCased space (last layer) is linearly
separable. So, the number of label pairs equals the number
of cluster pairs.} Based on an even split of the distances,
we partition all label pairs into three
bins: small, medium, and large. For each
task, we use the predictions of the best classifier to
compute the number of misclassified label pairs for each
bin. For example, if the clusters associated with the part of speech tags ADV and ADJ are close to each
other, and the best classifier misclassified ADV as ADJ, we
put this error pair into the bin of small distance. The
distribution of all errors is shown in
\Cref{tb:dis-error-rate}. This table shows that a large  majority of the
misclassified labels are concentrated in the small distance
bin. For example, in the supersense role task (SS-role),
$97.17\%$ of the errors happened in small distance bin.
The number of label pairs of each bin is shown in the
parentheses. \Cref{tb:dis-error-rate} shows that small
distances between clusters indeed confuse a classifier and
we can detect it \textit{without} training classifiers.

\begin{table}[]
\small
\centering
\begin{tabular}{lrrr}
\toprule
    \multirow{2}{*}{Task} & Small          & Medium        & Large       \\
                          & Distance       & Distance      & Distance    \\ \midrule
    SS-role               & 97.17\% (555)  & 2.83\% (392)  & 0\% (88)   \\
    SS-func               & 96.88\% (324)  & 3.12\% (401)  & 0\% (55)   \\
    POS                   & 99.19\% (102)  & 0.81\% (18)   & 0\% (16)    \\
    SR                    & 93.20\% (20)   & 6.80\% (20)   & 0\% (5) \\
    DEP                   & 99.97\% (928) & 0.03\% (103)  & 0\% (50)   \\ \bottomrule
\end{tabular} \caption{Error distribution based on different
    distance bins. The number of label pairs
in each bin is shown in the parentheses.}\label{tb:dis-error-rate}
\end{table}

\subsection{By-product: A Parameter-free
Classifier}\label{sec:by-product}

\begin{figure}[h]
   \centering
   \includegraphics[width=0.5\textwidth]{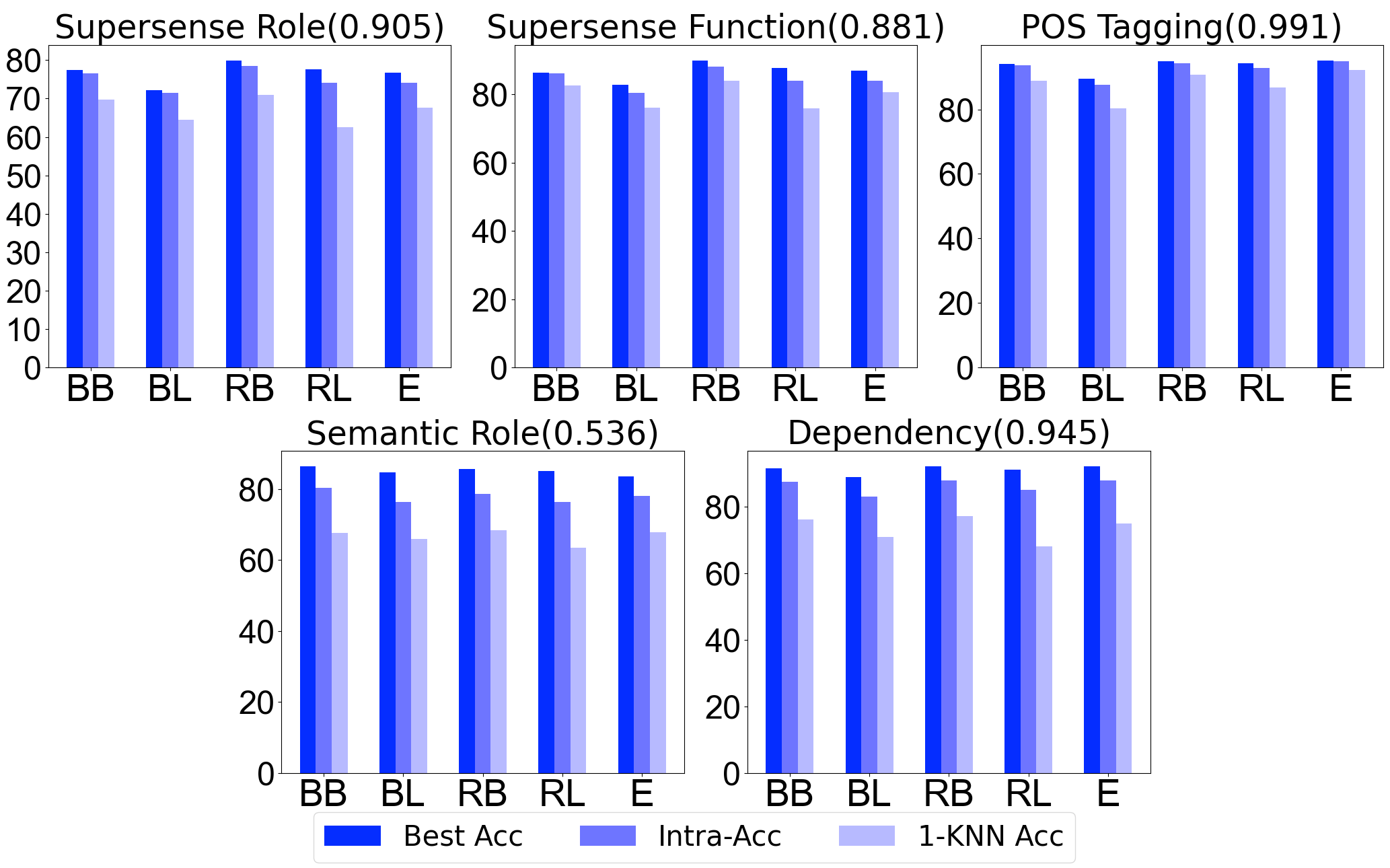}
   \caption{Comparison between the best classifier
   accuracy, intra-accuracy, and 1-kNN accuracy. The X-axis
    is different representation models. BB:
   \bertBaseCased, BL: \bertLargeCased, RB: \robertaBase,
   RL: \robertaLarge, E: \elmo. The pearson
   correlation coefficient between best classifier accuracy
   and intra-accuracy is shown in the parentheses
   alongside each task title. This figure is best viewed in color.}\label{fig:intra-acc}
\end{figure}

\paragraph{We can predict the expected performance of the
best classifier.}
Any $h\in V_{\epsilon}(\mathcal{H},E,D)$ is a predictor for the
task $D$ on the representation $E$. As a by-product of the
clusters from \probename, we can define a predictor. The prediction
strategy is simple: for a test example, we assign it to
its closest cluster.\footnote{To find the distance between the convex hull of a cluster and
a test point, we find a max-margin separating hyperplane by
training a linear SVM that separates the point from the
cluster. The distance is twice the distance between the
hyperplane and the test point.}
Indeed, if the label of the cluster is the
true label of the test point, then we know that there exists \emph{some}
classifier that allows this example to be correctly
labeled. We can verify the label every test point and compute the
aggregate accuracy to serve as an indicator of the
generalization ability of the representation at hand.  We
call this accuracy the \emph{intra-accuracy}. In other words, without
training classifiers, we can answer the question:
\emph{given a representation, what is the expected
performance of the best classifier for a task?}



\Cref{fig:intra-acc} compares the best
classifier accuracy, and the intra-accuracy
of the last layer of different embeddings.
Because our assignment strategy is similar to nearest neighbor classification (1-kNN), which assigns the unlabelled test point to its
closest labeled point, the figure also compares to the 1-kNN accuracy.

First, we observe that
intra-accuracy always outperforms the simple 1-kNN
classifier, showing that \probename can use
more information from the representation space. Second, we see
that the intra-accuracy is close to the best
accuracy for some tasks (Supersense tasks and POS tagging).
Moreover, all the pearson correlation coefficients between
best accuracy and intra-accuracy (showed in the parentheses
alongside each task title) suggest a high linear
correlation between best classifier accuracy and
intra-accuracy. That is, the intra-accuracy can be a good
predictor of the best classifier accuracy for a
representation. From this, we argue that
intra-accuracy can be interpreted as a benchmark accuracy of
a given representation without actually training classifiers. 

\subsection{Case Study: Identifying Difficult Examples}

The distances between a test point and all the clusters from the training set
can not only be used to predict the label but also can be used to identify
difficult examples as per a given representation. Doing so could lead to re-annotation of the data, and perhaps
lead to cleaner data, or to improved embeddings. Using the supersense role task,
we show a randomly chosen example of a mismatch between the annotated label and
the \bertBaseCased neighborhood:

\begin{quotation}
  \ldots \textbf{our} new mobile number is \ldots 
\end{quotation}

\newcommand{\gestalt}{\textsc{Gestalt}}
\newcommand{\possessor}{\textsc{Possessor}}

The data labels the word \textit{our} as \gestalt, while the embedding places it
in the neighborhood of \possessor. The annotation guidelines for these
labels~\cite{schneider2017adposition} notes that \gestalt\ is a supercategory of
\possessor. The latter is specifically used to identify cases where the
possessed item is alienable and has monetary value. From this definition, we see
that though the annotated label is \gestalt, it could arguably also be a
\possessor\ if phone numbers are construed as alienable possessions that have
monetary value. Importantly, it is unclear whether \bertBaseCased makes this distinction.
%
Other examples we examined required similarly nuanced analysis. This example
shows \probename can be used to identify examples in datasets that are
potentially mislabeled, or at least, require further discussion.


%% file: related.tex
\section{Related Work and Discussion}\label{sec:related}


In addition to the classifier based probes described in the rest of the paper,
a complementary line of work focuses on probing the representations using a
behavior-based methodology. Controlled test
sets~\cite{csenel2018semantic,jastrzebski2017evaluate} are designed and errors
are analyzed to reverse-engineer what information can be encoded by the
model~\cite[e.g.,][]{marvin-linzen-2018-targeted,DBLP:journals/corr/abs-2005-00719,wu-etal-2020-perturbed}.
Another line of work probes the space by ``opening up'' the representation space or
the model~\cite[e.g.,][]{DBLP:conf/nips/MichelLN19,voita-etal-2019-analyzing}.
There are some efforts to inspect the space from a geometric perspective~\cite[e.g.,][]{ethayarajh-2019-contextual,mimno-thompson-2017-strange}.
Our work extends this line of work to connect the geometric structure of
embedding space with classifier performance without actually training a
classifier.

Recent work~\cite{pimentel-etal-2020-information,voita-titov-2020-information,zhu-rudzicz-2020-information} probe
representations from an information theoretic perspective.
These efforts still need a probability distribution
$p(y\vert x)$ from a trained classifier. In
\Cref{sec:by-product}, we use clusters to predict labels. In the same vein, the conditional probability
$p(y\vert x)$ can be obtained by treating the negative
distances between the test point $x$ and all clusters as 
predicted scores and normalizing via softmax. Our
formalization can  fit into the information theoretic
analysis and yet avoid training a classifier. 

Our analysis and experiments open new directions for further research:

\noindent\textbf{Novel pre-training target:} The analysis presented here informs us that  larger distance between clusters
can improve classifiers. This could guide loss function design
when pre-training representations.

\noindent\textbf{Quality of a representation:} In this paper, we
focus on the accuracy of a representation. We could seek to measure other
properties (\eg, complexity) or proxies for them. These analytical approaches
can be applied to the $\epsilon$-version space to further analyze the quality of
the representation space.

\noindent\textbf{Theory of representation:} 
Learning theory, \eg VC-theory~\cite{vapnik2013nature}, describes the
learnability of classifiers; representation learning lacks of such
theoretical analysis. The ideas explored in this work ($\epsilon$-version
spaces, distances between clusters being critical) could serve as a foundation
for an analogous theory of representations.



%% file: conclusion.tex
\section{Conclusion}\label{sec:conclusion}

In this work, we ask the question: what makes a representation good
for a task? We answer it by developing \emph{\probename},  a heuristic
approach builds upon hierarchical clustering to approximate the
$\epsilon$-version space.  Via experiments with several
contextualized embeddings and linguistic tasks, we showed that \probename
can help us understand the geometry of the embedding space and ascertain when a representation can successfully be employed for a
task.


%% file: acknowledgments.tex
\subsection*{Acknowledgments}
\label{sec:acknowledgments}

We thank the members of the Utah NLP group and Nathan Schneider for discussions
and valuable insights, and reviewers for their helpful feedback. We also thank
the support of NSF grants \#1801446 (SATC) and \#1822877 (Cyberlearning).


%% file: appendix.tex
\section{\probename in Practice}\label{sec:detailed-algorithm}

In practice, we apply several techniques to speed up the
probing process. Firstly, we add two caching strategies:
\paragraph{Black List} Suppose we know two clusters $A$ and
$B$ can not be merged, then $A^{\prime}$ and $B^{\prime}$
\emph{cannot} be merged either if $A\subseteq A^{\prime}$ and
$B\subseteq B^{\prime}$ \paragraph{White List} Suppose we
know two clusters $A$ and $B$ can be merged, then
$A^{\prime}$ and $B^{\prime}$ \emph{can} be merged too if
$A^{\prime}\subseteq A$ and $B^{\prime}\subseteq B$

These two observations allow us to cache previous decisions in checking whether two clusters overlap. Applying these caching strategies can help us avoid
unnecessary checking for overlap, which is
time-consuming.

Secondly, instead of merging from the start to the end and
checking at every step, we directly keep merging to the end
without any overlap checking. After we arrived at the
end, we start checking backwards to see if there is an
overlap. If final clusters have overlaps, we find
the step of the first error, correct it and keep
merging. Merging to the end can also help us avoid plenty of
unnecessary checking because, in most representations, the
first error usually happens only in the final few merges.
\Cref{alg:practice} shows the whole algorithm.

\begin{algorithm}
    \small
    \caption{\probename in Practice}
    \begin{algorithmic}[1]
        \Require{A dataset $D$ with labeled examples $(x_i, y_i)$.}
        \Ensure{A set of clusters of points $\mathcal{C}$.}
        \State{Initialize $C_i$ as the cluster for each
        $(x_i,y_i)\in D$}
        \State{$\mathcal{C}=\{C_0, C_1, \cdots\}$}
        \State{Keep merging the closest pairs who have the
        same label in $\mathcal{C}$ without overlapping
        checking.}
        \If{There are overlappings in $\mathcal{C}$}
            \State{Find the step $k$ that first overlap happens.}
            \State{Rebuild $\mathcal{C}$ from step $k-1$}
            \State{Keeping merging the closest pairs who have the
                   same label in $\mathcal{C}$ with
                   overlapping checking.}
        \EndIf
        \State{\Return{$\mathcal{C}$}}
    \end{algorithmic}\label{alg:practice}
\end{algorithm}

\Cref{tb:running-time} shows the runtime comparison between
DirectProbe and training classifiers.

\begin{table}[]
\small
\begin{tabular}{@{}lrrrr@{}}
\toprule
        & \multicolumn{2}{c}{DirectProbe} & \multicolumn{2}{c}{Classifier} \\ 
        & Clustering       & Predict      & CV  & Train-Test \\\midrule
SS-role & 1 min            & 7 min        & 1.5 hours         & 1 hour     \\
SS-func & 1.5 min          & 5.5 min      & 1 hour            & 1 hour     \\
POS     & 14 min           & 43 min       & 3.5 hours         & 2 hours    \\
SR      & 10 min           & 22 min       & 50 min            & 1 hour     \\
DEP     & 3 hours          & 3 hours      & 20 hours          & 11 hours   \\ \bottomrule
\end{tabular}\caption{The comparison of running time between
    \probename and training classifiers using the setting
    described in \Cref{sec:classifier-details}. The time
    is computed on the \bertBaseCased space. \probename and
    classifiers run on the same machine.}\label{tb:running-time}
\end{table}

\section{Classifier Training Details}\label{sec:classifier-details}
We train $3$ different kinds of classifiers in order to find
the best one: Logistic Regression, a one-layer neural
network with hidden layer size of $(32, 64, 128, 256)$ and
two-layers neural network with hidden layer sizes of $(32,
64, 128, 256)\times (32, 64, 128, 256)$. All neural networks
use ReLU as the activation function and optimized by
Adam~\cite{kingma2014adam}. The cross-validation process
chooses the weight of each regularizer of each classifier.
The weight range is from $10^{-7}$ to $10^{0}$. We set the
maximum iterations to be $1000$. After choosing the best
hyperparameters, each specific classifier is trained $10$
times with different initializations.


%% file: details.tex
\section{Summary of Representations}\label{sec:rep-summarization}

Table~\ref{tb:representations} summarizes all the
representations in our experiments.

\begin{table}[]
\centering
\begin{tabular}{@{}lrr@{}}
\toprule
    Representations  & \#Parameters & Dimensions  \\ \midrule
    \bertBaseCased  & 110M         & 768        \\
    \bertLargeCased & 340M         & 1024       \\
    \robertaBase    & 125M         & 768        \\
    \robertaLarge   & 355M         & 1024       \\
    \elmo    & 93.6M        & 1024    \\ \bottomrule
\end{tabular}
\caption{Statistics of the five representations in our
    experiments.}\label{tb:representations}
\end{table}

\section{Summary of Tasks}\label{sec:task-summarization}
In this work, we conduct experiments on $5$ tasks, which
is designed to cover different usages of representations.
Table~\ref{tb:tasks} summarizes these tasks.

\begin{table*}[]
\centering
\begin{tabular}{@{}lrrccccc@{}}
\toprule
Task & \#Training & \#Test  & Token-based & Span-based & Pair-wise & Semantic & Syntax \\ \midrule
Supersense-role & \num{4282} & \num{457} & $\surd$ & & & $\surd$ & \\
Supersense-function & \num{4282} & 457 & $\surd$ & & & $\surd$ & \\
POS & \num{16860} & \num{4323} & $\surd$ & & & & $\surd$ \\
Dependency Relation  & \num{42081} & \num{4846} & & & $\surd$ & & $\surd$ \\ 
Semantic Relation  & \num{8000} & \num{2717} & & $\surd$ & $\surd$ & $\surd$ & \\ \bottomrule
\end{tabular}
\caption{Statistics of the five tasks with their different
    characteristics.}\label{tb:tasks}
\end{table*}

\section{Classifier Results v.s. \probename Results}\label{sec:detailed-probing}
Table~\ref{tb:probing-results} summarizes the
results of the best classifier for each representation and
task. The meanings of each column are as follows:
\begin{itemize}
    \item \textbf{Embedding}: The name of representation.
    \item \textbf{Best classification}: The accuracy of the best
        classifier among $21$ different classifiers. Each
        classifier run $10$ times with different starting
        points. The final accuracy is the average accuracy
        of these $10$ runs.
    \item \textbf{Intra-accuracy}: The prediction accuracy
        by assigning test examples to their closest clusters.
    \item \textbf{Type}: The type of the best classifier. \eg $(256,)$
        means one-layer neural network with hidden size
        $256$ and $(256,128)$ means two-layer neural network
        with hidden sizes $256$ and $128$ respectively.
    \item \textbf{Parameters}: The number of
        parameters of the best classifier.
    \item \textbf{Linear SVM}: \emph{Training} accuracy of
        a linear SVM classifier.
    \item \textbf{\#Clusters}: The number of final clusters
        after probing.
\end{itemize}

\section{Detailed Classification Results}\label{sec:cls}
\Cref{tb:cls-results} shows the detailed classification
results on Supersense-role task using the same settings
described in \Cref{sec:classifier-details}.
In this table, we record the difference between the minimum
and maximum test accuracy of the $10$ runs of each
classifier. The maximum difference for each representation
is highlighted by the underlines. The best
performance of each representation is highlighted by the
bold numbers. From this table, we observe:
\begin{inparaenum}[(i)]
\item the difference between different runs of the same
    classifier can be as large as 4-7\%, which can not be
    ignored;
\item different representation requires different model
    architecture to achieve its best performance.
\end{inparaenum}

\begin{table*}[]
\centering
\scriptsize
\begin{tabular}{@{}llrrrrrrrrrrrrrrr@{}}
\toprule
Model       & Specific  & \multicolumn{3}{c}{\bertBaseCased} & \multicolumn{3}{c}{\bertLargeCased} & \multicolumn{3}{c}{\robertaBase} & \multicolumn{3}{c}{\robertaLarge} & \multicolumn{3}{c}{\elmo} \\  
Family      &      Model                           & Min       & Max      & Diff       & Min       & Max       & Diff        & Min      & Max     & Diff      & Min       & Max      & Diff      & Min   & Max   & Diff    \\ \midrule
    Linear                             & LR              & 75.93     & 76.59    & 0.66             & 72.21     & 72.43     & 0.22              & 79.65    & 80.09   & 0.44            & 77.46     & 77.90    & 0.44            & 76.37 & \textbf{77.02} & 0.65          \\\midrule
    \multirow{4}{*}{One-layer}  & (32,)                           & 75.27     & 78.56    & 3.29             & 70.24     & 73.09     & 2.85              & 77.9     & 79.87   & 1.97            & 74.84     & \textbf{78.34}    & 3.5             & 74.4  & 75.93 & 1.53          \\
                                   & (64,)                           & 76.37     & 78.56    & 2.19             & 71.33     & 72.87     & 1.54              & 78.56    & 80.74   & 2.18            & 76.59     & 77.68    & 1.09            & 73.74 & 75.93 & 2.19          \\
                                   & (128,)                             & 75.71     & 77.9     & 2.19             & 70.46     & 73.3      & 2.84              & 77.9     & 80.53   & 2.63            & 75.71     & 77.46    & 1.75            & 73.52 & 76.15 & 2.63          \\
                                   & (256,)                             & 75.27     & 78.12    & 2.85             & 70.68     & \textbf{73.96}     & 3.28              & 78.34    & 80.96   & 2.62            & 74.84     & 77.9     & 3.06            & 73.96 & 76.81 & 2.85          \\\midrule
\multirow{16}{*}{Two-layers} & (32,32) & 71.99     & 76.81    & \underline{\textit{4.82}}    & 67.4      & 70.68     & 3.28              & 75.71    & 79.43   & 3.72            & 70.68     & 74.62    & 3.94            & 70.24 & 73.96 & 3.72          \\
    & (32,64)                           & 73.3      & 75.49    & 2.19             & 68.49     & 73.3      & 4.81              & 76.15    & 78.99   & 2.84            & 72.21     & 76.37    & 4.16            & 71.33 & 76.81 & \underline{\textit{5.48}} \\
                                   & (32,128)                          & 73.74     & 75.27    & 1.53             & 67.61     & 72.65     & 5.04              & 76.15    & 80.09   & 3.94            & 72.65     & 76.81    & 4.16            & 70.02 & 74.4  & 4.38          \\
                                   & (32,256)                          & 73.52     & 76.15    & 2.63             & 67.18     & 70.68     & 3.5               & 74.18    & 77.46   & 3.28            & 71.77     & 74.84    & 3.07            & 72.43 & 74.62 & 2.19          \\
                                   & (64,32)                           & 73.52     & 76.59    & 3.07             & 67.61     & 72.87     & 5.26              & 75.93    & 78.56   & 2.63            & 71.33     & 75.93    & \underline{\textit{4.6}}    & 71.99 & 75.49 & 3.5           \\
                                   & (64,64)                           & 72.87     & 76.59    & 3.72             & 68.05     & 71.77     & 3.72              & 76.81    & 79.21   & 2.4             & 73.74     & 75.49    & 1.75            & 73.52 & 75.49 & 1.97          \\
                                   & (64,128)                          & 73.09     & 76.81    & 3.72             & 69.37     & 72.43     & 3.06              & 76.81    & 79.21   & 2.4             & 72.65     & 75.49    & 2.84            & 72.87 & 75.27 & 2.4           \\
                                   & (64,256)                          & 73.96     & 76.59    & 2.63             & 68.49     & 71.55     & 3.06              & 76.59    & 78.77   & 2.18            & 73.52     & 76.37    & 2.85            & 73.3  & 76.59 & 3.29          \\
                                   & (128,32)                          & 72.65     & 77.24    & 4.59             & 70.02     & 72.43     & 2.41              & 76.59    & 80.53   & \underline{\textit{3.94}}   & 73.74     & 75.93    & 2.19            & 72.43 & 75.93 & 3.5           \\
                                   & (128,64)                          & 73.74     & 77.02    & 3.28             & 68.93     & 73.3      & 4.37              & 76.59    & 79.87   & 3.28            & 72.65     & 75.93    & 3.28            & 72.21 & 76.37 & 4.16          \\
                                   & (128,128)                         & 74.84     & 77.24    & 2.4              & 69.37     & 72.43     & 3.06              & 77.46    & 79.21   & 1.75            & 72.87     & 75.71    & 2.84            & 73.52 & 76.15 & 2.63          \\
                                   & (128,256)                        & 73.74     & 77.24    & 3.5              & 64.33     & 72.21     & \underline{\textit{7.88}}     & 77.02    & 80.74   & 3.72            & 72.87     & 76.15    & 3.28            & 73.09 & 76.59 & 3.5           \\
                                   & (256,32)                          & 73.74     & 78.12    & 4.38             & 68.05     & 73.3      & 5.25              & 77.68    & 80.31   & 2.63            & 73.74     & 75.93    & 2.19            & 72.87 & 75.71 & 2.84          \\
                                   & (256,64)                          & 75.49     & \textbf{78.77}    & 3.28             & 68.93     & 72.43     & 3.5               & 77.68    & 80.09   & 2.41            & 73.74     & 75.71    & 1.97            & 72.87 & 76.15 & 3.28          \\
                                   & (256,128)                         & 75.27     & 78.77    & 3.5              & 70.46     & 73.96     & 3.5               & 77.9     & 80.74   & 2.84            & 73.74     & 77.02    & 3.28            & 73.96 & 76.37 & 2.41          \\
        & (256,256) & 75.71     & 78.34    & 2.63             & 66.74 & 73.52     & 6.78 & 78.12    & \textbf{81.18}   & 3.06            & 74.84 & 76.37    & 1.53 & 73.3  & 75.71 & 2.41 \\ \bottomrule
\end{tabular}\caption{Classification results on Supersense role task. Each
specific classifier is trained ten times with different initializations. We
record the minimum and maximum performance of these ten runs.  Diff is the
difference between the minimum and maximum performance. Bold
number highlights the best performance of each
representation. Underlines highlights the maximum difference
between ten runs. See \Cref{sec:cls} for a discussion.}\label{tb:cls-results}
\end{table*}

\begin{table*}[]
\centering
\small
\begin{tabular}{@{}llrrlrrr@{}}
\toprule
Task & Embedding & Best classification & Intra-accuracy &
    Type & Parameters & Linear SVM & \#cluster \\ \midrule
\multirow{5}{*}{SS-role} & \bertBaseCased & $77.48\pm
    0.92$ & 76.58 & (256,128) & \num{235264} & 100 & 46 \\
& \bertLargeCased & $72.25\pm 0.09$ & 71.55 & linear & \num{47104} & 100 & 46 \\
& \robertaBase & $79.85\pm 0.15$ & 78.56 & linear & \num{35328} & 100 & 46 \\
    & \robertaLarge & $77.7\pm 0.18$ & 74.18 & linear & \num{47104} & 100 & 46 \\
& \elmo  & $76.7\pm 0.24$  & 74.18 & linear & \num{47104} & 100 & 46 \\
\midrule
\multirow{5}{*}{SS-func} & \bertBaseCased & $86.3\pm
    0.44$ & 86.21 & (128,) & \num{103424} & 100      & 40        \\
& \bertLargeCased & $82.84\pm 0.73$ & 80.53 & (256,) & \num{272384} & 100 & 40 \\
& \robertaBase & $89.87\pm 0.51$ & 88.18 & (256,) & \num{206848} & 100 & 40  \\
& \robertaLarge & $87.72\pm 0.56$ & 84.03 & (256,128) & \num{300032} & 100 & 40 \\
& \elmo & $86.87\pm 0.31$ & 84.03  & linear & \num{40960} & 100 & 40 \\
\midrule
\multirow{5}{*}{POS} & \bertBaseCased & $94.11\pm 0.14$ & 93.59 & (256,256) & \num{266496} & 100  & 17\\
& \bertLargeCased & $89.54\pm 0.34$ & 87.69 & (128,256) &\num{168192} & 100 & 17 \\
& \robertaBase & $95\pm 0.09$ & 94.26 & (128,256) & \num{135424} & 100 & 17\\
& \robertaLarge & $94.25\pm 0.19$ & 92.92 & (128,) & \num{133248} & 99.97 & 1487\\
& \elmo  & $95.08\pm 0.1$ & 94.93 & (256,) & \num{266496} & 100 & 17\\
\midrule
\multirow{5}{*}{Semantic Relation} & \bertBaseCased & $86.43\pm 0.28$ & 80.27 & (256,) & \num{395776} & 100 & 10\\
& \bertLargeCased & $84.71\pm 0.27$ & 76.41 & (128,256) & \num{297472} & 100 & 10\\
& \robertaBase & $85.55\pm 0.28$ & 78.58 & (256,32)  & \num{401728} & 100 & 10\\
& \robertaLarge & $85.04\pm 0.23$ & 76.26 & (256,) & \num{526848} & 100 & 10 \\
& \elmo & $83.47\pm 0.28$ & 78.06 & (128,128) & \num{279808} & 100 & 10 \\
\midrule
\multirow{5}{*}{Dependency} & \bertBaseCased & $91.52\pm 0.25$ & 87.49 & (256,) & \num{405248} & 100 & 47 \\
& \bertLargeCased & $88.9\pm 0.38$ & 83.06 & (256,) & \num{536320} & 100 & 47 \\
& \robertaBase & $92.24\pm 0.03$ & 87.99 & linear & \num{72192} & 100 & 47 \\
& \robertaLarge & $91.18\pm 0.04$ & 85.06 & linear & \num{96256} & 100 & 47\\
& \elmo & $92.21\pm 0.25$ & 88.05 & (256,) & \num{536320} & 100 & 47 \\
\bottomrule
\end{tabular}
\caption{Intra-accuracy results on $5$ tasks, compared against the best
  classifier results. See \Cref{sec:detailed-probing} for details about each column.}\label{tb:probing-results}
\end{table*}
